\documentclass[letterpaper]{article} 
\usepackage{aaai24}  
\usepackage{times}  
\usepackage{helvet}  
\usepackage{courier}  
\usepackage[hyphens]{url}  
\usepackage{graphicx} 
\usepackage{natbib}  
\usepackage{array}
\usepackage{caption} 
\usepackage{algorithm}
\usepackage{algorithmic}
\usepackage{amsmath}
\usepackage{amssymb}
\usepackage{newfloat}
\usepackage{listings}
\DeclareCaptionStyle{ruled}{labelfont=normalfont,labelsep=colon,strut=off} 
\floatstyle{ruled}
\newfloat{listing}{tb}{lst}{}
\floatname{listing}{Listing}
\usepackage{multirow}
\usepackage{booktabs}
\usepackage{color}

\usepackage[capitalize]{cleveref}
\crefname{section}{Sec.}{Secs.}
\Crefname{section}{Section}{Sections}
\Crefname{table}{Table}{Tables}
\crefname{table}{Tab.}{Tabs.}

\title{Learning from History: Task-agnostic Model Contrastive Learning \\ for Image Restoration}
 
\author{
    Gang Wu, 
    Junjun Jiang\thanks{Corresponding Author},
    Kui Jiang,
    Xianming Liu
}
\affiliations{
      Faculty of Computing, Harbin Institute of Technology, Harbin 150001, China \\
\{gwu,~jiangjunjun,~jiangkui,~csxm\}@hit.edu.cn
}

\usepackage{bibentry}

\begin{document}

\maketitle

\begin{abstract}
Contrastive learning has emerged as a prevailing paradigm for high-level vision tasks, which, by introducing properly negative samples, has also been exploited for low-level vision tasks to achieve a compact optimization space to account for their ill-posed nature. However, existing methods rely on manually predefined and task-oriented negatives, which often exhibit pronounced task-specific biases. To address this challenge, our paper introduces an innovative method termed 'learning from history', which dynamically generates negative samples from the target model itself. Our approach, named Model Contrastive Learning for Image Restoration (MCLIR), rejuvenates latency models as negative models, making it compatible with diverse image restoration tasks. We propose the Self-Prior guided Negative loss (SPN) to enable it. This approach significantly enhances existing models when retrained with the proposed model contrastive paradigm. The results show significant improvements in image restoration across various tasks and architectures. For example, models retrained with SPN outperform the original FFANet and DehazeFormer by 3.41 and 0.57 dB on the RESIDE indoor dataset for image dehazing. Similarly, they achieve notable improvements of 0.47 dB on SPA-Data over IDT for image deraining and 0.12 dB on Manga109 for a 4$\times$ scale super-resolution over lightweight SwinIR, respectively. Code and retrained models are available at https://github.com/Aitical/MCLIR.

\end{abstract}

\section{Introduction}

Image restoration, aiming at recovering a high-quality image from the degraded one, is a fundamental problem in the fields of image processing and computer vision \cite{survey_derain_from_model_to_data,survey_sisr,survey_face}. Deep learning approaches have made considerable advancements in image restoration, while there are still challenges due to its ill-posed nature \cite{CNN-survey,survey_IR_2022}. The success of the self-supervised learning paradigm for high-level tasks, especially those using contrastive learning methods, has drawn great attention \cite{survey_contrastive_2023}. This inspires many researchers to make strides in improving the end-to-end learning paradigm for image restoration tasks, incorporating the concept of sample contrastive learning and bridging the gap between high-level and low-level tasks \cite{contrastive_dehazing_compact,contrastive_derain,contrastive_derain_self_simlarity,contrastive_sisr_PCL,contrastive_dehazing_cvpr23}. Image restoration tasks usually contain high-quality ground truth as the learning target (positive sample), and more attention is paid to obtaining appropriate negative examples. For example, Wu \textit{ et al.}~\cite{contrastive_dehazing_compact} directly utilized low-quality input as the negative sample and introduced the contrastive paradigm for the image dehazing task. Wu \textit{et al.} proposed a hard negative construction for image super-resolution tasks. Compared to the super-resolved results, hard negatives with similar image quality to the anchor sample can push it ahead effectively \cite{contrastive_sisr_PCL}. Most recently, Zheng \textit{et al.} utilized multiple pre-trained models to provide consensual negatives and proposed a progressively improved negative lower bound for image dehazing \cite{contrastive_dehazing_cvpr23}.

\begin{figure}[!t]
    \centering
    \includegraphics[width=0.45\textwidth]{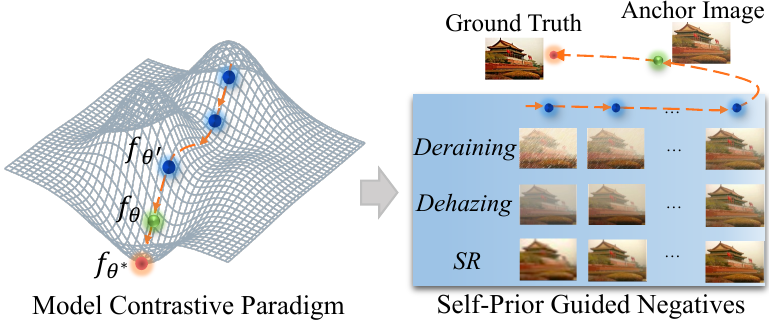}
    \caption{
    Illustration of the proposed model contrastive paradigm. We provide a common optimization space for it. To the target model $f_{\theta}$, the proposed model contrastive paradigm exploits negative samples from the latency model $f_{\theta^{'}}$ smoothly. Compared to task-oriented negatives in previous work, our model contrastive paradigm is task-agonist and general to various image restoration tasks. This provides a compact optimization space adaptively (pushing target model $f_{\theta}$ closer to assumed optimal $f_{\theta^{*}}$).} \label{fig:toy_optimization} 

\end{figure}

While existing contrastive learning methods for low-level tasks, leveraging potent approaches such as hard negative mining~\cite{contrastive_sisr_PCL} and curriculum learning strategies \cite{contrastive_dehazing_cvpr23}, have shown impressive performance, certain intrinsic limitations persist. Chiefly, there exists an \textit{over-reliance on task-oriented prior}, leading to limited generalization capability across multiple image restoration tasks. Many existing methods exhibit a pronounced task-oriented bias, where negative sample generation is often influenced by prior knowledge and empirical evaluation centered on a singular target task \cite{contrastive_dehazing_compact,contrastive_sisr_PCL,contrastive_dehazing_cvpr23}. Such approaches, while effective in isolation, often impede to a diverse set of image restoration tasks and model architectures. In light of these challenges, it compels us to ask: \textit{Is there a task-agnostic and general method for negative sampling that could potentially enhance the performance of a diverse range of image restoration tasks?}

Taking this into consideration, we turn our focus to the target model itself, rather than sample selection. In recent literature, much of the spotlight has been directed towards negative sample collection, often overlooking a latent gem, the latency model, during the learning process. Specifically, the latency model, when operating within a close optimization step, shares strikingly similar parameters with the current model. This intrinsic similarity paves the way for the construction of proper negatives pertinent to the current anchor sample. To illustrate this concept, we introduce a toy example in Fig.~\ref{fig:toy_optimization}. It becomes evident that, throughout the learning journey, the output of the latency model exhibits a suboptimal but congruent distribution relative to the current anchor sample. This alignment offers the potential to derive 'hard' negatives that are well suited to the task at hand. Furthermore, as the entire learning process is incrementally refined, the negatives in our model contrastive paradigm adopt a curriculum way naturally. Motivated by these insights, we put forth an innovative \textit{\textbf{model contrastive paradigm}} for image restoration tasks. In contrast to previous approaches with complex negative mining strategies, the core of our proposed model contrastive paradigm lies in the self-promoted negatives between latency and current models. Notably, it is a task-agonist and general to diverse image restoration tasks.

\begin{figure*}[!ht]
    \centering    \includegraphics[width=\textwidth]{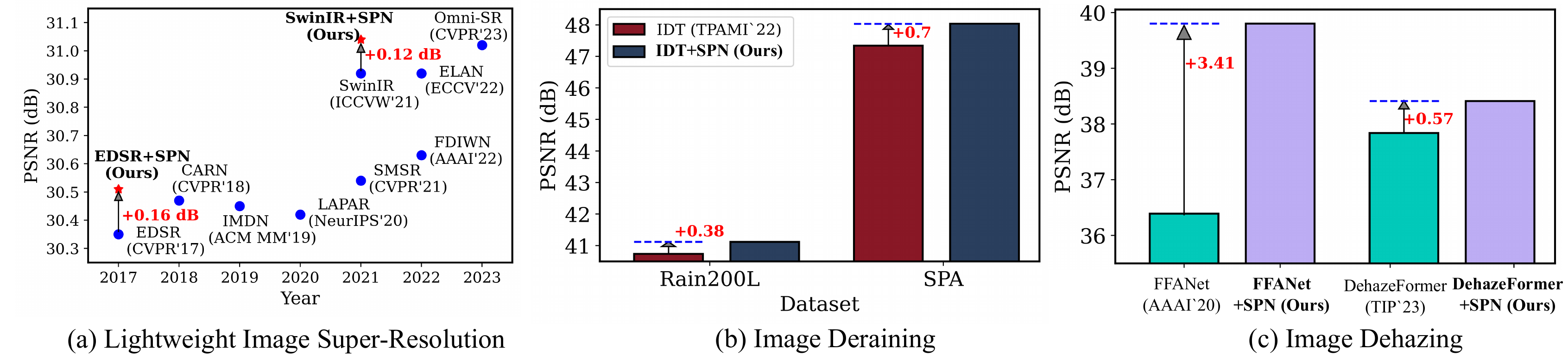}
    \caption{Comparisons between models retrained by our proposed model contrastive paradigm and the originals. Retrained models can achieve remarkable improvements on various image restoration tasks.}
    \label{fig:demo_improvement}
\end{figure*}

In detail, we propose a model contrastive paradigm for various image restoration tasks (MCLIR) with a Self-Prior guided Negative loss (SPN). Importantly, the proposed MCLIR is compatible with existing approaches across different architectures and tasks. We retrain existing models by the proposed model contrastive and some results are presented in Fig.~\ref{fig:demo_improvement}. Retrained models can be found to achieve promising improvement across multiple tasks and architectures. More specifically, based on lightweight EDSR \cite{EDSR} and SwinIR \cite{SwinIR}, our retrained models show superior performance on the Manga109 test dataset for $\times4$ scale image super resolution, boasting improvements of \textbf{ 0.16 dB}  and \textbf{ 0.12 dB} in terms of PSNR. Retrained IDT \cite{derain_IDT} achieves \textbf{{0.38 dB}} and \textbf{{0.7 dB}} gains on Rain200L and SPA datasets for image deraining. Furthermore, for image dehazing, we gain a notable improvement of \textbf{ 3.41 dB} and \textbf{ 0.57 dB} compared to original FFANet \cite{FFANet} and DehazeFormer \cite{DehazeFormer}, respectively.

The main contributions of this work are:
\begin{itemize}
\item The paper proposes a novel approach MCLIR through a task-agnostic model contrastive paradigm, which provides the adaptive generation of negative samples directly from the target model itself. Unlike conventional methods that manually apply negative samples to a specific target task, the proposed model contrastive paradigm exhibits versatility across multiple tasks and models.

\item The paper introduces the self-prior guided negative loss (SPN) for image restoration, which is seamlessly compatible with existing methods. SPN provides a simple to enhance existing image restoration models by integrating self-supervision principles within our model contrastive paradigm.

\item The paper demonstrates the effectiveness of the proposed approach by retraining existing models with the MCLIR, which significantly improves image restoration across various tasks and architectures. 
\end{itemize}

\section{Related Work}

\subsubsection{Image Restoration} Image restoration, crucial in image processing and computer vision,  aims to restore high-quality images from degraded versions \cite{survey_image_restoration_neuralcomputing}. Deep learning, particularly CNNs, has revolutionized this domain \cite{survey_sisr,survey_face,survet_derain,survey_dehaze}, with advancements in various CNN-based methods \cite{SRCNN,DNCNN,DCNNIR_zhangkai,DCNNDerain}, and the incorporation of novel architectures such as residual networks \cite{EDSR}, attention mechanisms \cite{RCAN,Attention_Dehaze}, and UNet structures \cite{UNet_CNN,MSPFNet_jiang}. As well as some lightweight architectures were proposed for practical application \cite{real_time_derain,real_time_low_light,IMDN,wu2023fully,ESRT,wu2023incorporating}. Recent trends include Transformer-based architectures \cite{SwinT,ViT}, delivering breakthroughs in many image restoration tasks \cite{SwinIR,DehazeFormer,Restormer,derain_IDT,Uformer,low_light_transformer}. In this work, we turn our attention to the optimizing strategy and exploit an effective model contrastive learning method to refresh the performance existing models.

\subsubsection{Contrastive Learning} Self-supervised learning, particularly contrastive methods, has seen remarkable success in high-level vision tasks \cite{survey_contrastive_2023}. In the field of low-level image restoration, researchers are exploring the integration of self-supervised regularization to bridge the gap between low- and high-level tasks \cite{contrastive_blind_degradation,contrastive_dehazing_compact,contrastive_derain,contrastive_derain_self_simlarity,contrastive_dehazing_cvpr23,contrastive_sisr_PCL}. Typically, low-level tasks often have high-quality ground truth, making the identification of informative negative samples vital. Wu \textit{et al.} employed low-quality inputs as negative samples and introduced a perceptual-based contrastive loss for more effective convergence \cite{contrastive_dehazing_compact}. Wu \textit{et al.} proposed a practical contrastive learning framework for single image super-resolution tasks, enhancing performance through hard negative construction and negative information interpolation \cite{contrastive_sisr_PCL}. Nonetheless, the adaptability and generalization of these methods across diverse image restoration models and tasks remain a challenge.

Our work distinguishes itself from existing methods \cite{contrastive_dehazing_compact,contrastive_sisr_PCL,contrastive_dehazing_cvpr23} by introducing a task-agnostic model contrastive paradigm for image restoration. This novel approach shifts the focus from negative samples to negative models, offering a straightforward yet effective framework for a variety of image restoration tasks.

\section{Method}

\subsection{Overall Framework}

Deep learning-based image restoration methodologies have recently made considerable breakthroughs across multiple restoration tasks, including image super-resolution, image dehazing, image deblurring, image deraining, and so on. In this paper, we focus our attention on the vanilla end-to-end framework with deep learning methods.

Given a low-quality input image $I^{LQ}$, the target model $f_{\theta}$ (where $\theta$ denotes the model parameters) processes this input to produce a reconstructed image $I^{Rec}$. The high-quality counterpart of the image is represented as $I^{HQ}$. The optimization of $f_{\theta}$ is guided by minimizing the reconstruction loss $\mathcal{L}_{rec}$, defined as:
\begin{equation}
    \min_{\theta} \mathcal{L}_{rec}( f_{\theta}(I^{LQ}), I^{HQ}),
\end{equation}
where $\mathcal{L}_{rec}$ typically depends on metrics such as Mean Absolute Error (MAE) and Mean Squared Error (MSE).

 Delving further, existing contrastive learning methods for image restoration \cite{contrastive_dehazing_compact,contrastive_dehazing_cvpr23,contrastive_sisr_PCL} utilize negative samples $I^{Neg}$, extending the standard end-to-end learning with a negative loss $\mathcal{L}_{neg}$:
\begin{equation}
    \min_{\theta} \mathcal{L}_{rec}(f_{\theta}(I^{LQ}), I^{HQ}) + \mathcal{L}_{neg}(f_{\theta}(I^{LQ}), I^{Neg}),
\end{equation}
where the negative samples are predefined and generated based on task-specific priors, which limit their generalization capability across diverse image restoration tasks.

In this work, we introduce a novel model contrastive paradigm for image restoration, shifting from task-specific negative samples to negatives derived from the target model itself. We designate the target model as $f_{\theta}$, representing its state in a given training iteration $t$. A key innovation is the development of a latency model, $f_{\theta^{'}}$ (the negative model), which is utilized to produce adaptive negative samples for training $f_{\theta}$. This approach simplifies the overall process in existing methods and provides a general framework for image restoration. The specifics of implementing negative models and the corresponding loss functions will be elaborated in subsequent sections.

\subsection{Learning from History}

We propose a practical implementation for leveraging latency models to generate informative negative samples during training, as introduced earlier. This method avoids the challenges of significant gaps between model checkpoints and the impracticality of frequent processing, especially with larger models. In detail, we introduce the efficient strategy of exponential moving averages (EMA) to achieve a smooth negative model. The update equation for the negative model is as follows:

\begin{equation}
 \theta^{'} = w\theta^{'} +(1-w)\theta,~ \mathrm{ s.t. }~ t \% s = 0.
\label{eq:ema}
\end{equation}

In this schema, $w$ denotes the update weight, $s$ is the update step, and $t$ captures the current iteration. To ensure the preservation of latency parameters, we adopt a long step $s$ and selectively update the negative model $f_{\theta}$ at intervals of every $s$ iteration.

\subsection{Self-Prior Guided Negative Loss}
At the heart of our model contrastive paradigm, a loss function mediates between the target reconstruction $I^{Rec}$ and its negative counterpart, $I^{Neg}$. Here, we take the pre-trained VGG \cite{VGG} network as the embedding network to map the samples into a latent feature space, where $f^{Rec}=\text{VGG}(I^{Rec})$ and $f^{Neg}=\text{VGG}(I^{Neg})$. Then the proposed negative loss $\mathcal{L}_{neg}$ is formulated as

\begin{equation}
\mathcal{L}_{neg} = \|f^{Rec}-f^{Neg}\|_{1}.
\end{equation}

Furthermore, our model-based contrastive paradigm can incorporate multiple negatives by adding more negative models. 
For more robustness, we take multiple distinct steps to obtain several 
latency models. The combined negative loss, accounting for multiple negatives, is represented as
\begin{equation}
    \mathcal{L}^{N}_{neg}=\frac{1}{N}\sum_{i=1}^{N}{\|f^{Rec}-f^{Neg}_{i}\|}_{1},
\end{equation}
where $N$ denotes the total number of negative models and $f_{i}^{Neg}$ corresponds to the latent feature of the $i$-th negative sample.

Compared to existing methods, our negative regularization is a self-prior guided loss function, where the negative samples stem from the target model itself. More important, it is general and transferable to existing image restoration models while retaining the original learning strategy. Typically, image restoration tasks utilize a reconstruction loss, $\mathcal{L}{rec}$, relying on metrics such as Mean Absolute Error (MAE) and Mean Squared Error (MSE). To validate the prowess of our model-based contrastive paradigm by $\mathcal{L}_{neg}$, we have retrained numerous exiting methods with it, testing across different image restoration tasks and architectures. The precise formulation of the reconstruction loss $\mathcal{L}_{rec}$ is dependent on the retraining method.

Generally, the total loss function within our model-based contrastive paradigm is defined as:

\begin{equation}
\mathcal{L} = \mathcal{L}_{rec}-\lambda \mathcal{L}^{N}_{neg},
\end{equation}
where $\mathcal{L}_{rec}$ represents the corresponding reconstruction loss adopted in existing method, and $\lambda$ is the balancing coefficient. This simple formulation allows us to incorporate the proposed SPN with existing image restoration methods, enhancing their flexibility and adaptability to various tasks.

\begin{table*}[!tbp]
\renewcommand\arraystretch{1.1}
\centering
\resizebox{0.95\textwidth}{!}{
\begin{tabular}{|cccccccc|}
\hline
Methods & Architecture  & Scale & \begin{tabular}[c]{@{}c@{}}Avg.\\ PSNR/SSIM \end{tabular} & \begin{tabular}[c]{@{}c@{}}Set14\\ PSNR/SSIM\end{tabular}& \begin{tabular}[c]{@{}c@{}}B100\\ PSNR/SSIM\end{tabular}& \begin{tabular}[c]{@{}c@{}}Urban100\\ PSNR/SSIM\end{tabular} & \begin{tabular}[c]{@{}c@{}}Manga109\\ PSNR/SSIM\end{tabular} \\ 
\hline
\hline

EDSR-light &  \multirow{4}{*}{\hfil CNN}  & $\times2$ 
& 34.06/0.9303 
& 33.57/0.9175 
& 32.16/0.8994 
& 31.98/0.9272 
& 38.54/0.9769\\

\textbf{+SPN (Ours)} & &  
& \textbf{34.19/0.9313}	 
& \textbf{33.67/0.9182} 
& 	\textbf{32.21/0.9001}	 
& {\textbf{32.23/0.9297}}	 
& {\textbf{38.64/0.9772}} \\

EDSR-light &  & $\times4$& 28.14/0.8021
& 28.58/0.7813
& 27.57/0.7357 
& 26.04/0.7849
& 30.35/0.9067  \\

\textbf{+SPN (Ours)} & &  & \textbf{28.21/0.8040}	 & \textbf{28.63/0.7829} & 	\textbf{27.59/0.7369}	 & {\textbf{26.12/0.7878}}	 & {\textbf{30.51/0.9085}}\\

\hline

SwinIR-light & \multirow{4}{*}{\hfil Transformer}  & $\times4$
& 28.46/0.8099
& 28.77/0.7858 
&27.69/0.7406 
&26.47/0.7980  
&30.92/0.9151 \\

\textbf{+SPN (Ours)} &  &
& \textbf{28.55/0.8114}
&\textbf{28.85/0.7874}
&\textbf{27.72/0.7414} 
&{\textbf{26.57/0.8010}} 
&{\textbf{31.04/0.9158}} \\

SwinIR &  & $\times4$
& 28.88/0.8190
& 28.94/0.7914
&27.83/0.7459
&27.07/0.8164
&31.67/0.9226 \\

\textbf{+SPN (Ours)} &  &   
& \textbf{28.93/0.8198}
&\textbf{29.01/0.7923}
&\textbf{27.85/0.7465} 
&{\textbf{27.14/0.8176}}    
&{\textbf{31.75/0.9229}} \\

\hline
\end{tabular}
}
\caption{{Comparison results of image super-resolution.} We take CNN-based EDSR~\cite{EDSR} and Transformer-based SwinIR \cite{SwinIR} as our baselines. Results of retrained models by the proposed model contrastive paradigm are in {bold}. Avg. presents the mean value of four test datasets.} \label{tab:edc_table} 
\end{table*}

\begin{figure*}[!ht]
    \centering
    \includegraphics[width=0.95\textwidth]{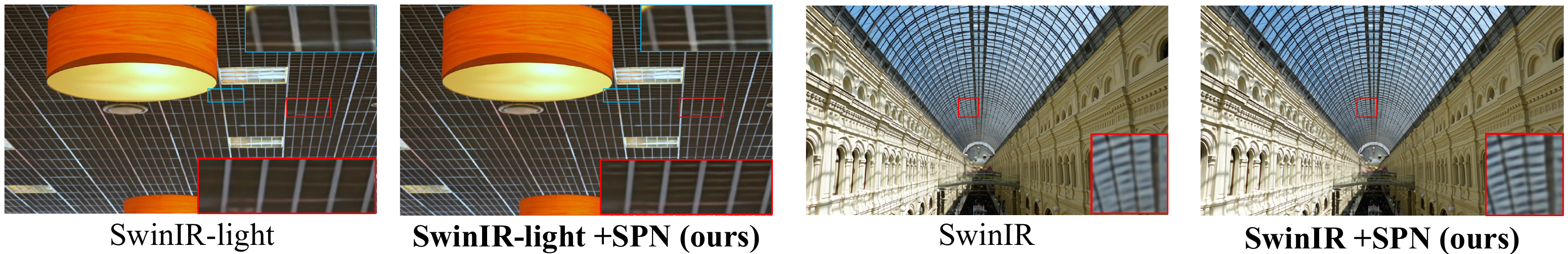}
    \caption{Visual comparison for image super-resolution tasks. Displayed are results from both the original SwinIR models and those retrained by our model contrastive paradigm. The enhancements brought about by our approach are clearly evident. } \label{fig:vision_sr}
    
\end{figure*}

\subsection{Remark}
In this study, we propose the model contrastive paradigm for image restoration tasks. This approach significantly simplifies the construction of negatives while simultaneously providing adaptive and effective ones. Contrary to existing approaches \cite{contrastive_dehazing_compact,contrastive_sisr_PCL,contrastive_dehazing_cvpr23}, the proposed model contrastive paradigm does not rely on the restrictions of task-oriented priors, making it versatile and universal to various image restoration tasks. In essence, our contribution lies in providing a task-agonist and general approach to the construction of negative samples, expanding the scope of contrastive learning in the field of image restoration.

\section{Experiment}

\subsection{Experimental Settings}
In this section, we apply the following four image restoration tasks to test and validate the effectiveness of the proposed method.

\subsubsection{Image Super-Resolution}
We retrain the CNN-based EDSR \cite{EDSR} and Transformer-based SwinIR \cite{SwinIR}. We take 800 images from DIV2K \cite{DIV2K} for training and test them on five benchmark datasets. 
\subsubsection{Image Dehazing}
We employ the CNN-based FFANet \cite{FFANet} and  Transformer-based DehazeFormer \cite{DehazeFormer} as our baselines and retrain them for image dehazing tasks.  Following \cite{DehazeFormer}, we utilize the indoor training dataset (ITS) and RESIDE-6K to separately train the models and evaluate on the synthetic objective testing set (SOTS).

\subsubsection{Image Deblurring}
For image deblurring, we retrain and evaluate NAFNet \cite{NAFNet} on GoPro dataset \cite{nah2017deep}.

\subsubsection{Image Deraining}
For image deraining, we conduct experiments on several publicly available benchmarks, namely Rain200L/H \cite{data_rain200}, DID-Data \cite{data_DID}, DDN-Data \cite{data_ddn}, and SPA-Data \cite{data_SPA}. We adopt the Transformer-based IDT model \cite{derain_IDT} as our baseline and subsequently retrain it using the proposed model contrastive paradigm. 

Given the diverse training settings across various image restoration tasks, we ensure a fair comparison by strictly adhering to the original training configurations of each retrained model, as specified in their literature.

\subsection{Comparison Results}

\begin{table*}[h]
\renewcommand\arraystretch{1.1}
\centering

\scalebox{0.95}{
\begin{tabular}{|lcccccc|}
\hline
Methods  & 
\begin{tabular}[c]{@{}c@{}} Avg.\\ PSNR/SSIM\end{tabular} &
\begin{tabular}[c]{@{}c@{}}Rain200L\\ PSNR/SSIM\end{tabular} &
\begin{tabular}[c]{@{}c@{}}Rain200H\\ PSNR/SSIM\end{tabular} &
\begin{tabular}[c]{@{}c@{}}DID\\ PSNR/SSIM\end{tabular} &
\begin{tabular}[c]{@{}c@{}}DDN\\ PSNR/SSIM\end{tabular} & \begin{tabular}[c]{@{}c@{}}SPA\\ PSNR/SSIM\end{tabular} \\

\hline \hline
(CVPR21) MPRNet & 36.17/0.9543 & 39.47/0.9825 & 30.67/0.9110 & 33.99/0.9590 & 33.10/0.9347 & 43.64/0.9844 \\
(AAAI'21) DualGCN &36.69/0.9604 &40.73/0.9886 & 31.15/0.9125 & 34.37/0.9620 & 33.01/0.9489 & 44.18/0.9902 \\
(ICCV'21) SPDNet &36.54/0.9594 & 40.50/0.9875 & 31.28/0.9207 & 34.57/0.9560 & 33.15/0.9457 & 43.20/0.9871 \\
(CVPR'22) Uformer-S & 36.95/0.9505& 40.20/0.9860& 30.80/0.9105   & 34.46/0.9333 & 33.14/0.9312 & 46.13/0.9913 \\
(CVPR'22) Restormer  & 37.49/0.9530 & 40.58/0.9872 & 31.39/0.9164 & 35.20/0.9363 & 34.04/0.9340 & 46.25/0.9911\\
(CVPR'23) DRSformer& 38.33/0.9676 & 41.23/0.9894 & 32.17/0.9326  &35.35/0.9646 & 34.35/0.9588 & 48.54/0.9924 \\

\hline
(TPAMI'22) IDT
&37.77/0.9593
&40.74/0.9884
&32.10/0.9343
&34.85/0.9401
&33.80/0.9407	
&47.34/0.9929 \\

\textbf{(Ours) IDT+SPN} 
& \textbf{38.03/0.9610}
& \textbf{41.12/0.9893}
& \textbf{32.17/0.9352}

&\textbf{34.94/0.9424} 
&{\textbf{33.90/0.9442}} 
&{\textbf{48.04/0.9938}} \\

\hline
\end{tabular}
} 
\caption{{Comparison results of image deraining.} We take IDT \cite{derain_IDT} as the benchmark and retrain it with the proposed model contrastive paradigm. We evaluate the performance on several image deraining datasets, and our results are in {bold} (Average performance of the five datasets is calculated in the Avg. column). }\label{tab:result_derain}
\end{table*}

\begin{figure*}[!ht]
    \centering
    \includegraphics[width=0.95\textwidth]{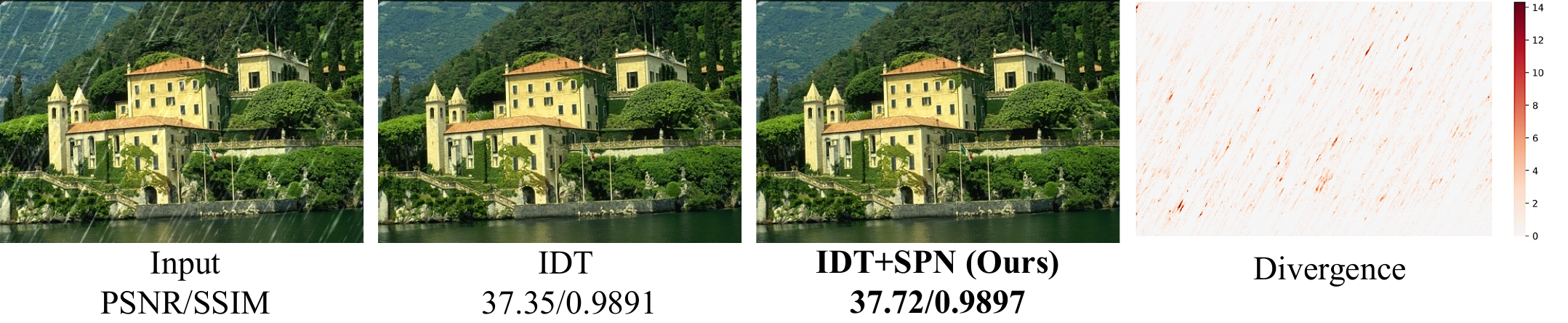}
    \caption{Visual comparisons between IDT and our retrained one. A divergence map delineates the differences between the IDT output and ours, highlighting the improvement, particularly in degraded regions. }\label{fig:vision_deraining}
    
\end{figure*}

\subsubsection{Image Super-Resolution}
We first investigate the image super-resolution task, utilizing both CNN-based EDSR and Transformer-based SwinIR as baselines to evaluate our proposed self-prior guided negative regularization. Results presented in Tab.~\ref{tab:edc_table} across various architectures and model capacities.  Our proposed model contrastive paradigm can achieve considerable improvements when retraining lightweight models such as EDSR and SwinIR, achieving average gains of 0.07 dB and 0.09 dB on scale $\times$4, respectively. Moreover, the retrained EDSR model demonstrates a noteworthy average improvement of 1.3 dB for the scale $\times$2 task. In particular, there are substantial improvements from our re-training on the Urban100 and Manga109 datasets. This implies that our model contrastive paradigm facilitates a more accurate and compact convergence in the restoration process. Moreover, comparisons with current state-of-the-art methods are presented in Fig.~\ref{fig:demo_improvement} (a). Remarkably, the retrained EDSR and SwinIR can eclipse the performance of several contemporary methods. These underscore the promise of our model contrastive paradigm.

In addition to the lightweight models, we also retrain the large SwinIR model. Our proposed approach exhibits its general applicability to larger models, providing an average improvement of 0.05 dB. It is worth noting that the improvements seen in the larger models were relatively smaller compared to the lightweight models. This observation is consistent with the understanding that lightweight models, owing to their smaller capacity, may struggle to find an optimal solution. Therefore, our proposed model contrastive paradigm provides significant assistance in these cases. Conversely, larger models have the capacity to fit or even overfit the dataset, hence the model contrastive paradigm improves them within a lesser margin. Visual results are showcased in Fig.~\ref{fig:vision_sr}. It is also evident that our model contrastive paradigm can improve existing methods, particularly in bringing more precise textures.

\subsubsection{Image Deraining}
We take the advanced IDT \cite{derain_IDT} as a baseline and retrain it by the proposed model contrastive paradigm on five datasets, and Tab.~\ref{tab:result_derain} provides a comprehensive comparison. The retrained IDT achieves an average PSNR improvement of 0.26 dB across all test datasets. For instance, it showcases a gain of 0.7 dB in comparison to its original in the SPA dataset. Furthermore, visual results are illustrated in Figure~\ref{fig:vision_deraining}, accompanied by a divergence map. It intuitively highlights enhancements in the degraded regions. These insights underscore the ability of our model contrastive paradigm to refine target degradation.

\begin{table}[!t]
 \renewcommand\arraystretch{1.25}
  \centering
  
    {
\resizebox{0.475\textwidth}{!}{
      \begin{tabular}{|l|cc|cc|}
        \hline
        \multirow{3}*{Methods}&\multicolumn{2}{c|}{ITS}&\multicolumn{2}{c|}{RESIDE-6K}\\
        \cline{2-5}
         &\multicolumn{2}{c|}{SOTS-indoor}&\multicolumn{2}{c|}{SOTS-mix} \\
        \cline{2-5} & PSNR  & SSIM    & PSNR  & SSIM \\
        \hline
        \hline
        (CVPR'21) AECR-Net  & 37.17 & 0.990 & - & -  \\
        (ICLR'23) ~SFNet & 41.24 & 0.996 & -& - \\
        (CVPR'23) C$^2$PNet & 42.56 &0.995 & - & - \\
        \hline
        (AAAI'20) FFANet  & 36.39 & 0.989 &  29.96 & 0.973 \\
        \textbf{(Ours) FFANet+SPN }        
        &\textbf{{39.80}} &  \textbf{0.995} & \textbf{30.65} & \textbf{0.976} \\
        \hline
       (TIP'23) DehazeFormer-T   & 35.15 & 0.989 & {30.36} & 0.973 \\
        \textbf{(Ours) DehazeFormer-T+SPN }  & \textbf{35.51} & \textbf{0.990} & {\textbf{30.44}} & \textbf{0.974}\\
       (TIP'23)  DehazeFormer-S  & 36.82 & {0.992} &  {30.62} & {0.976}\\
        \textbf{(Ours) DehazeFormer-S+SPN}  & \textbf{37.24} & \textbf{0.993} & \textbf{{30.77}}& \textbf{0.978} \\
         (TIP'23) DehazeFormer-B & {37.84} & {0.994} & {31.45} & 0.980 \\
        \textbf{(Ours) DehazeFormer-B+SPN }  & \textbf{38.41}& \textbf{0.994}   &\textbf{31.57 }& \textbf{0.981} \\
        \hline
      \end{tabular}
    }
}\caption{ {Comparison results of image dehazing.} The results of our retrained models are in bold}. 
  \label{tab:result_dehaze} 
\end{table}

\subsubsection{Image Dehazing}
\begin{figure}[!ht]
    \centering
    \includegraphics[width=0.475\textwidth]{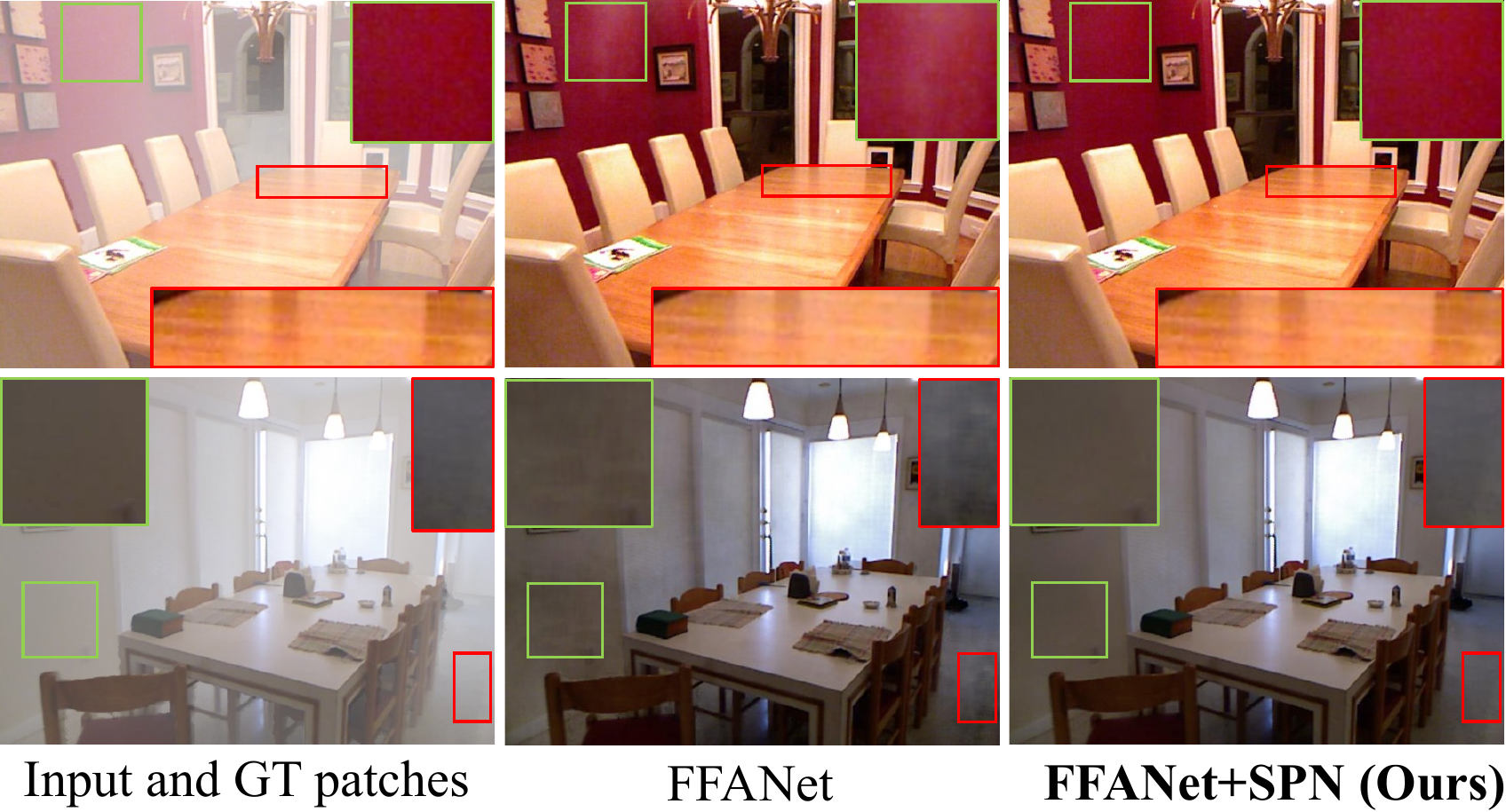}
    \caption{Visual results of FFANet and our retrained one for image dehazing. }\label{fig:vision_ffanet}
\end{figure}

Following the recent work~\cite{DehazeFormer}, we utilize the SOTS-indoor and SOTS-mix datasets for testing. In detail, we retrain FFANet \cite{qin2020ffa} and DehazeFormer \cite{DehazeFormer}, with the results detailed in Fig.~\ref{tab:result_dehaze}. From the table, a remarkable enhancement is achieved over FFANet, exhibiting gains of 3.41 dB and 0.69 dB on the indoor and mixed test datasets, respectively. Concurrently, the Transformer-based model DehazeFormer also records advancements, evident across various model scales. Specifically, the retrained DehazeFormer-B achieves the most pronounced improvement of 0.57 dB on the indoor test dataset. Intriguingly, the retrained FFANet even outperforms the DehazeFormer. It is reasonable that FFANet is a CNN-based model, which has a large model capacity with heavy parameter counts, and our model contrastive paradigm appears to steer FFANet towards more optimized results. Some visual results are presented in Fig.~\ref{fig:vision_ffanet}. One can find that the retrained FFANet is clearer with fewer artifacts. 

\subsubsection{Image Deblurring}
Results of the retrained NAFNet \cite{NAFNet} for image deblurring are in Tab.~\ref{tab:result_deblur}. One can find that the retrained NAFNet showcases notable enhancements compared to the original and outperforms the Transformer-based Restormer \cite{Restormer}.

\subsubsection{Comparison to Existing Sample Contrastive Paradigm}
In contrast to existing methods that employ task-oriented negatives, our model contrastive paradigm provides straightforward enhancement to image restoration tasks. We furnish a comprehensive comparison between exiting contrastive approaches and our innovative model contrastive paradigm. For the image dehazing task, we examine previous methods based on FFANet, including contrastive regularization (CR) \cite{contrastive_dehazing_compact}, curricular contrastive regularization (CCR) \cite{contrastive_dehazing_cvpr23}, and our self-prior guided negative loss (SPN). The comparative results can be found in Tab.~\ref{tab:contrastive_loss}. Instead of merely employing low-quality input images as negative samples, CCR achieves significant advancements by utilizing consensus negatives from pre-trained models and the curriculum learning approach. Intriguingly, our SPN outperforms other methods, achieving the best improvement without additional prior knowledge. In addition, for image super-resolution \cite{contrastive_sisr_PCL}, our model contrastive paradigm again registers the highest gains, as presented in Tab.~\ref{tab:contrastive_loss}. This consistent performance underscores the fact that our proposed model contrastive paradigm is task-agnostic and highly effective for diverse image restoration tasks.

\begin{table*}[h]
\renewcommand\arraystretch{1.1}
\centering
\small{
\begin{tabular}{|c|ccccc|cc|}
\hline Methods  & MIMO-UNet & HINet & MAXIM  &Restormer  & UFormer & NAFNet & \textbf{NAFNet+SPN (Ours)} \\
\hline
\hline PSNR & 32.68 & 32.71 & 32.86 & 32.92 & 32.97 & 32.87 & \textbf{32.93}\\
SSIM & 0.959 & 0.959 & 0.961 & 0.961 & 0.967  &  0.9606 & \textbf{0.9619} \\
\hline
\end{tabular}
}
\caption{{Comparison results of image deblurring.}  We take NAFNet \cite{NAFNet} as the benchmark and retrain it with the proposed model contrastive paradigm on GoPro dataset.}\label{tab:result_deblur}
\end{table*}

\begin{table}[!ht]
  \centering

\renewcommand\arraystretch{1.15}{
\small
{
\begin{tabular}{|c|c|cc|}
        \hline
        Methods & {Task \& Dataset} & PSNR  & SSIM\\
        \hline
        \hline
        FFANet &    & 36.39 & 0.9886 \\
        +CR & Image Dehazing & 36.74 & 0.9906 \\
        +CCR & (SOTS-indoor )& 39.24 & 0.9937 \\
        \textbf{+SPN (Ours)} && \textbf{39.80} & \textbf{0.9947}\\
        \hline
        EDSR & SISR & 26.04 & 0.7849 \\
        +PCL &  (Urban100)& 26.07 & 0.7863 \\
        \textbf{+SPN (Ours)} && \textbf{26.12} & \textbf{0.7878}\\
        \hline
      \end{tabular}
    }  
}
    \caption{A comparative analysis of the existing contrastive paradigms versus our proposed model contrastive approach. Typically, existing methods are task-oriented and are proposed for image dehazing (CR \cite{contrastive_dehazing_compact} and CCR \cite{contrastive_dehazing_cvpr23}) and image super-resolution (PCL \cite{contrastive_sisr_PCL}), separately. {Our model contrastive paradigm is task-agonist and outperform existing methods}.
  }\label{tab:contrastive_loss}

\end{table}

\begin{table}[!t]
\renewcommand\arraystretch{1.2}
\centering
\resizebox{0.475\textwidth}{!}{
\begin{tabular}{|c|cccc|}
\hline
Negative Model & - & Random &  Pre-trained & Default \\
\hline\hline
Avg. PSNR  & 28.46  & 28.44 & 28.51 & 28.55 \\
\hline
\end{tabular}
}
\caption{Results of retrained lightweight SwinIR with different negative models. }\label{tab:ablation_nagetive}
\end{table}

\subsection{Ablation Studies}
\subsubsection{Impact of Negative Model}
This ablation study aims to elucidate the role of our negative model by investigating various configurations, including a randomly initialized and fixed SwinIR model, a pre-trained and fixed SwinIR model, and our standard model. The results, detailed in Tab.~\ref{tab:ablation_nagetive}, highlight the effectiveness of our approach. The baseline model is denoted by '-,' indicating the origin SwinIR. We found that a negative model with random and fixed parameters hinders the target SR model, failing to provide effective negatives. Conversely, using a pre-trained and fixed negative model shows improvement, but not as much as our model contrastive. This suggests the importance of adaptive and curriculum-based negative generation in our model contrastive paradigm, aligned with the findings of existing methods \cite{contrastive_dehazing_cvpr23}.

\begin{table}[!t]
\renewcommand\arraystretch{1.25}
\centering

\resizebox{0.4\textwidth}{!}{
\begin{tabular}{|l|ccccc|}
\hline 
Step $s$ & 100 &  500 & 1000 & 2000 & All \\
\hline \hline
FFANet & 38.90  & 38.54 & 38.27 & 37.55 & \textbf{39.80} \\
\hline
\end{tabular}
}
\caption{Ablation studies of the negative step $s$. We retrain FFANet with different negative step $s$ on indoor dataset. }\label{tab:ablation_step}
\end{table}

\subsubsection{Impact of Negative Step}
We conducted an ablation study, in which FFANet was retrained with various negative models, each differentiated by its update iteration, $s$. The results are tabulated in Tab.~\ref{tab:ablation_step}. It becomes evident that the choice of step $s$ distinctly influences both the negative model and performance. A smaller step provides more challenging negatives compared to the larger step, leading to superior performance. In our experiments, we employ multiple negative models with the four steps together, achieving the best results. The incorporation of multiple negative models ensures a consistent supply of robust negatives.

\subsubsection{Impact of Balancing Coefficient}
We study the influence of the coefficient $\lambda$ in Eq.~(\ref{eq:ema}), and results are presented in Tab.~\ref{tab:ablation_lambda}, where the value 0 means the original loss $L_1$ without our negative regularization. When $\lambda$ is set to 1e-2 or 1e-3, there is collapse during training, especially in the early stage. This is reasonable because large negative loss can influence the model updating towards the optimal direction. When $\lambda$ is 1e-4, our model contrastive paradigm achieves the best performance. Considering that the proposed model contrastive paradigm is general to various image restoration tasks, in this paper we take $\lambda=1e-4$ as the default value across different tasks.

\begin{table}[!t]
\renewcommand\arraystretch{1.2}
\centering
\resizebox{0.475\textwidth}{!}{
\begin{tabular}{|c|cccccc|}
\hline
$\lambda$ & 0 &  1e-2 & 1e-3 & 5e-4 & 1e-4 & 1e-5  \\
\hline
\hline
EDSR-light & 28.58 & - &  - & 28.56 & 28.63 & 28.60 \\
\hline
\end{tabular}
}
\caption{Ablation studies on coefficient in total loss function. }\label{tab:ablation_lambda}
\end{table}

\begin{table}[!t]
\renewcommand\arraystretch{1.25}
\centering

\resizebox{0.425\textwidth}{!}{
\begin{tabular}{|l|cccccc|}
\hline
$w$ in EMA & 0 & 0.01 & 0.1 & 0.5 & 0.9  & 0.999 \\
\hline\hline
EDSR-light & - & - & \textbf{28.63} & 28.60 & 26.60 & 28.61 \\
\hline
\end{tabular}
}
\caption{Ablation studies on the updating weight in EMA. }\label{tab:ablation_ema}
\end{table}

\subsubsection{Impact of Updating Weight in EMA}
In reference to Eq.~(\ref{eq:ema}), we examine the influence of varying updating weights. The outcomes are detailed in Tab.~\ref{tab:ablation_ema}. In general, a smaller value of $w$ retains less latency information, bringing it in closer alignment with the target model. This results in the generation of the most challenging negatives. The symbol '-' indicates model collapse during training. As $w$ increases, it becomes evident that the optimal performance is attained at $w=0.1$. On the basis of our empirical observations, we consistently use $w=0.1$ as the default setting in all subsequent experiments.

\subsection{Discussion and Limitation}
In this paper, we proposed a novel model contrastive paradigm for low-level image restoration tasks, improving existing models. Although our experiments cover a range of tasks, areas such as JPEG artifact removal, image denoising, and certain real-world scenarios remain unexplored. Our ablation studies focused on image super-resolution, as shown in Tab.~\ref{tab:ablation_nagetive} and ~\ref{tab:ablation_step} but we did not extensively evaluate our hyperparameters across all tasks. These unexamined aspects offer valuable opportunities for future research.

\section{Conclusion}
In this study, we propose an innovative model contrastive paradigm for various low-level tasks. Compared to the task-oriented negatives in existing methods, the proposed model contrastive paradigm, constructing negatives from the target model itself, is task-agonist and general to various image restoration tasks by a self-prior guided negative loss (SPN). Our proposed SPN is straightforward to implement. We have retrained several image restoration models, and they achieve significant improvements across various tasks and architectures. In the future, we believe it would be meaningful to evaluate our proposed paradigm in more dense prediction tasks, potentially offering fresh insights and advances for the community.

\section{Acknowledgments}
The research was supported by the National Natural Science Foundation of China (U23B2009, 92270116).

\bibliography{main}

\end{document}